\documentclass[journal]{IEEEtran}
\usepackage{amsmath}
\usepackage{tabularx}
\usepackage{url,lineno,microtype}
\usepackage{float}
\usepackage{placeins}

\usepackage[inline]{enumitem}
\usepackage{subcaption}
\usepackage{booktabs}
\usepackage{graphicx}
\usepackage{xcolor}
\usepackage{color, colortbl}
\usepackage{multirow}
\usepackage[hidelinks,breaklinks=true]{hyperref}
\usepackage{layouts}
\usepackage[backend=biber,
    bibstyle=ieee,              
    citestyle=numeric-comp,     
    sortcites=true,   
    giveninits=true,            
    doi=false,
    maxbibnames=3
]{biblatex}
\usepackage[official]{eurosym}
\definecolor{Gray}{gray}{0.9}
\bibliography{bibliography.bib}
\graphicspath{{./Figures/}}
\DeclareGraphicsExtensions{.pdf,.png,.jpeg,.jpg}
\title{\LARGE \bf An Open-Source Modular Treadmill for Dynamic Force Measurement with Load Dependant Range Adjustment}
\author{Alborz {Aghamaleki Sarvestani}$^{1,2}$, Felix Ruppert$^{1,2}$, and Alexander Badri-Spr\"owitz$^{2,3}$\\%
\thanks{$^{1}$ The authors contributed equally to the paper}%
\thanks{$^{2}$Dynamic Locomotion Group, Max Planck Institute for Intelligent Systems, 70569 Stuttgart, Germany, $^{3}$Department of Mechanical Engineering, KU Leuven, 3001 Leuven, Belgium}%
}
\begin{document}
\maketitle
\begin{abstract}
Ground reaction force sensing is one of the key components of gait analysis in legged locomotion research. To measure continuous force data during locomotion, we present a novel compound instrumented treadmill design. The treadmill is 1.7\,m long, with a natural frequency of 170\,Hz and an adjustable range that can be used for humans and small robots alike. Here, we present the treadmill's design methodology and characterize it in its natural frequency, noise behavior and real-life performance.  Additionally, we apply an ISO 376 norm conform calibration procedure for all spatial force directions and center of pressure position. We achieve a force accuracy of $\leq 5.6$\,N for the ground reaction forces and $\leq 13$\,mm in center of pressure position.
\end{abstract}
\section{Introduction}
\begin{figure*}
	\centering
  \includegraphics[width=.75\textwidth]{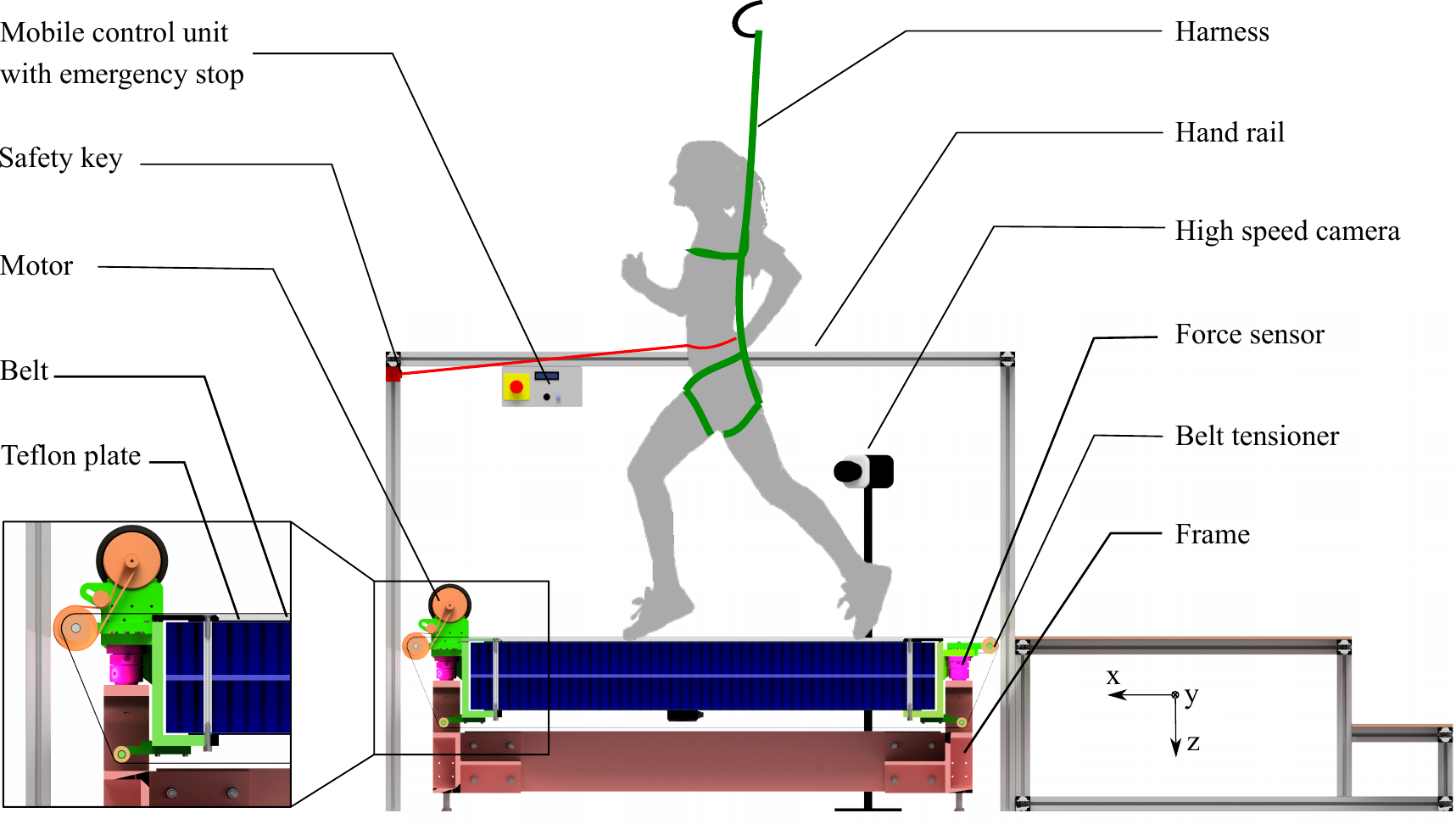}
	\caption{Cut view of the treadmill render with base frame (red), force sensors (purple), a compound plate fixture (green), actuation (orange), compound plate consisting of a sandwich plate (dark blue), carbon fiber plate (light blue) handrail (grey) and mobile control unit with emergency stop and treadmill safety key (red) on top. The compound plate, fixture, and actuator are floating on the force sensors to ensure that all internal forces are measured. The handrail is not connected to the treadmill to not compromise the natural frequency.}
	\label{fig:instrumentedTreadmillCut}
\end{figure*}
In biomechanics and locomotion research, one keystone data is ground reaction force data \cite{Daley2006}. Ground reaction force (GRF) data enables locomotion research in humans and animals \cite{Daley2006,Schumacher2018}, where no internal measurement of dynamics is possible. In robotic research, ground reaction forces can be used to validate bio-inspired leg designs \cite{Ruppert2019,Sproewitz2013,ashtiani2021hybrid,badri2022birdbot} as well as control designs \cite{Semini2015,Hubicki2016}.\\
The gold standard sensors for locomotion research are stationary force plates \cite{Bates1983}. While force plates are a great tool to measure the forces exerted onto the floor during a step, their major drawback is that they can only capture one step due to their stationary nature and small size. Solutions to this problem are either wearable sensor solutions \cite{Chuah2014, Ruppert2020} or force plates with moving substrate \cite{Kram1998}.\\
By incorporating force sensors into a treadmill, continuous data can be captured. With a split-belt treadmill, where each foot's step is recorded by an individual treadmill, gaits with a double-stance phase (e.g., walking) can be characterized \cite{Willems2013}. Even treadmills with three belts can be realized \cite{Paolini2007}.\\
The driving factor in the design of force sensors is the natural frequency of the sensor as well as the whole measuring system. The high natural frequency is required so that the treadmill does not oscillate in the frequency spectrum examined in gait analysis, usually $\approx$ 30\,Hz (\cite{PamiesVila2012}). This is especially important when measuring subjects with a low body mass like rodents or small robots \cite{Witte2002}. To adapt to  different requirements for lightweight and heavy subjects, treadmills of different sizes can be implemented (\ref{tab:treadmillComparison}). However, both small- and large-sized treadmills are prized about the same (with the highest costs being the sensors), which leads to twice the overall price.\\
Previously reported instrumented treadmills can be divided into these categories; commonly, treadmill force sensors are built from strain gauge sensors \cite{Bundle2015, Dierick2004, Verkerke2005}. While strain gauge sensors are a good solution to measure static loads, piezo-based sensors are more suited for measuring dynamic load under high static preload \cite{Belli2001}. While piezo sensors are more expensive, they are stiffer because the deflection measured in the force transducer is on an atomic level. Due to their higher natural frequency, the higher stiffness enables piezo sensors to measure extremely fast signals above 1 kHz.  \\
A simpler solution is to place a recreational treadmill on top of factory-made force plates \cite{Collins2009, Kram1998}. While the effort in design and production is small, store-bought treadmills with plywood surfaces can not provide the stiffness and rigidity needed to measure forces with high temporal solutions.\\
A general design decision is the placement of components with respect to the force sensors. The motor can be placed outside the treadmill and connected with a flexible drive shaft to minimize motor vibrations in the sensor signal and to reduce weight \cite{Bundle2015}. Alternatively, the motor can be placed on top of the force sensors, so the sensors are the only connection of the treadmill to the ground \cite{Collins2009, Belli2001}. While an external motor reduces weight, its parasitic connection to the ground can degrade the signal quality since not all forces are recorded by force sensors.\\

In addition to ground reaction forces, the center of pressure (COP) is an essential measurement in legged locomotion research \cite{Benda1994,Sloot2015, Goldberg2009}. The COP describes the point where the GRF vector applies on the foot \cite{Benda1994}. To calculate the COP from the force vectors of the individual sensors, an accurate measuring system is required. Both the force data, as well as the dimensions of the treadmill have to have high accuracy to achieve accurate COP estimations \cite{PamiesVila2012}.\\
To calibrate the instrumented treadmill, several different methods have been reported. A straightforward way is to use calibrated dead weights to compare the sensor output \cite{Bundle2015, Kram1998, Paolini2007, Verkerke2005}. While this method is simple to implement and norm conform, it only allows calibration in the vertical direction. To calibrate all three force directions, an instrumented stick with a force sensor can be used \cite{Collins2009}. With this method, the application of forces in three dimensions is possible. Still, since the stick is pushed manually, the results are not reproducible, which is mandatory in norm conform calibration. The calibration results are, therefore, not comparable between different treadmill designs.\\
In this paper, we present an open-source, modular, instrumented treadmill with a compound plate. We first provide a framework to estimate the natural frequency of a similar system for different geometric design parameters. To achieve maximum stiffness at reduced weight, we designed a novel compound surface structure made from aluminium and carbon fiber. We then quantify the treadmill, showing improved natural frequency, noise behavior, and accuracy in force and center of pressure sensing. We also present a norm conform protocol to calibrate treadmills based on international norms and standards. Last we showcase the ability of our treadmill to measure ground reaction forces over a wide measuring range. We evaluate the treadmill performance for a small robot of 4\,kg as well as a human subject of 90\,kg body weight.
The novelty of this work is fourfold:
\begin{itemize}
\item We present a design methodology for achieving high natural frequencies based only on the dimensions of the surface area. We validate our approach and achieve a natural frequency of only 4\% off the model calculation. The high required stiffness is achieved through a compound surface structure made from an aluminium honeycomb plate and carbon fiber plates for increased tensile strength.
\item We present a modular, compound, instrumented treadmill design. The high natural frequency with adjustable range enables the use for heavy subjects (90\,kg human) as well as for small, light subjects (5\,kg robot)
\item We present a calibration routine that conforms to ISO 376 and DKD 3-3 to achieve accurate and reproducible calibration results that are not possible with previously presented approaches. Our calibration approach allows a standardized comparison of different treadmill designs.
\item We present a calibration method to increase the COP measurement on the treadmill through a reproducible calibration that increases the COP accuracy to less than 1\,mm. We showcase the importance of this calibration approach in scenarios where the sensor plane and the surface plane are not in the same plane.
\end{itemize}

\begin{table*}
\centering
\caption{Comparison of key features for different instrumented treadmills presented in the literature.}
\label{tab:treadmillComparison}
\renewcommand{\arraystretch}{1.3} 
\newcolumntype{g}{>{\columncolor{Gray}}l}
\begin{tabularx}{\textwidth}{@{}l l l l l l l g @{}}
\toprule
Parameter & Belli \cite{Belli2001} & Bundle \cite{Bundle2015} & Collins \cite{Collins2009} & Kram \cite{Kram1998} & Paolini \cite{Paolini2007} & Verkerke \cite{Verkerke2005} & A. Sarvestani et. al.\\
\midrule
Natural frequency [Hz]&58\,Hz&113\,Hz &41\,Hz &87\,Hz &210\,Hz &120\,Hz &170\,Hz\\
Sensor type & Piezo & Strain gauge& Strain gauge& Strain gauge  & - & Strain gauge & Piezo\\
Sample rate [Hz]& 800 & 2000 & 1200 & 1000 & 1080 & 100 & 10\,000\\
Noise & 5\,N at 17\,Hz &-& 50\,N at 40\,Hz & 80\,N at 46\,Hz & 5.1\,N std. dev. & -  & 6\,N at 70\,Hz\\
Floating mass [kg] & 250 & - & 140 & 90 & - & - & 45\\
Calibration & Instrumented  & Weight & Instrumented  & Weight & Weight & Weight & ISO 376, DKD 3-9 \\
& leg & & pole & & & &in all spatial directions  \\
\bottomrule
\end{tabularx}
\end{table*}
\section{Tools and Methods}
\subsection{Mechanical Design}
The basic concept of this instrumented treadmill (\autoref{fig:instrumentedTreadmillCut}) is to float all moving parts that are in contact with the treadmill surface on top of the force sensors. In this design, all forces from the actuator and the measuring forces are recorded. The treadmill is constructed so that two treadmills can be placed side by side to form a split-belt treadmill with a belt gap of 44\,mm.
The treadmill base consists of a frame made from steel and longitudinal aluminium profiles. The frame acts as the fixture point for the force sensors. Additionally, the heavy frame acts as a counterweight to the ground reaction forces exerted on the treadmill during locomotion. The surfaces in contact with the force sensors are precision ground to ensure a plane connection between the sensor and the frame. The force sensors sit on top of the base frame and connect the treadmill compound plate and the compound fixture to the base frame.\\
To achieve maximum stiffness at low mass, the compound plate consists of two honeycomb sandwich plates (10\,mm thickness, 35$\mu$m honeycomb thickness) and two carbon fiber plates bonded with an epoxy resin for aluminium-carbon connections (2015B, \textit{Araldite}). The compound plate is clamped into two c-shaped compound fixtures made from aluminium, including the connection for the force sensors. To decouple the whole treadmill assembly from the frame, the actuator is also placed on the compound fixture to ensure that all internal forces are captured in the sensors. The actuator is connected to a standard conveyor belt running over four rollers (ROFAWT50-15-L400, \textit{Misumi}). The top rollers are barreled to ensure self-centering of the belt. The surface of the treadmill in contact with the underside of the conveyor belt is covered with a Teflon plate to reduce gliding friction between the belt and the treadmill. The roller in the back of the treadmill is equipped with a custom-built tensioning mechanism based on a lead screw to adjust the belt tension to remove belt slippage on the rollers.\\
    The motor is a brushed DC treadmill motor taken from a commercial treadmill (\textit{Christopeit-Sport}). It has 0.92\, kW continuous power at 180\,V bus voltage and a maximum motor speed of 5400\,rpm. The motor is geared with a 3:1 ribbed V-belt (PJ, 483x5\,mm, \textit{zahnriemen24} and is connected to a motor controller (DPCANIE-015S400, \textit{a-m-c}). The ribbed V-belt was chosen because the timing belt in the initial design introduced vibrations into the system. Because of the undercut between the small motor-side timing pulley (8 teeth) and the belt, an unacceptable noise source at a frequency of $f_{noise} = \omega_{motor}\cdot n_{teeth}$ was visible in the Fourier transform of the force signal. \\
The secondary motor axis is equipped with a flywheel to increase inertia and stabilize motor speed. Motor position is measured with a magnetic angular encoder (AEAT8800, \textit{Broadcom}) and fed back to the motor controller. To use the treadmill at a constant speed, the motor controller can be set to speed control mode. The motor controller also has a torque control mode with direct access to the desired current. In this mode, it is possible to run perturbation experiments with reproducible speed perturbations.\\
The force sensors (9366CCSP, \textit{Kistler}) are connected to a summation box under the compound plate. The summation box is connected to a charge amplifier (9865, \textit{Kistler}) and a data acquisition system (85695B, \textit{Kistler}). The charge amplifier can automatically switch its amplification stage depending on the number of charges coming from the sensors. The sensors have sensitivities of $\approx$ 7.7\,$\frac{pC}{N}$ in x- and y-direction and $\approx$ 3.9\,$\frac{pC}{N}$ in z-direction. The system can therefore be used for loads ranging from 129 - 6493\,N in x- and y-direction and 256 - 12820\,N in z-direction(\autoref{tab:ranges}).
\begin{table}
\centering
\caption{Maximum force ranges and resolutions for different amplifier settings}
\label{tab:ranges}
\renewcommand{\arraystretch}{1.3}
\begin{tabularx}{\columnwidth}{@{}l r l r l @{}}
\toprule
 Amplifier  & Force range & Resolution &Force range & Resolution \\
 range [pC] & x,y [N]& x,y [N] & z [N] & z [N]\\
\midrule
 1000  & 129.9  & 0.0019 & 256.4   & 0.0039\\
 5000  & 649.4  & 0.0099 & 1282.1  & 0.0195\\
 10\,000 & 1298.7 & 0.0198 & 2564.1  & 0.0391\\
 50\,000 & 6493.5 & 0.0991 & 12820.5 & 0.1956\\
 \bottomrule
\end{tabularx}
\end{table}
\begin{table}
\centering
\renewcommand{\arraystretch}{1.3}
\caption{Instrumented Treadmill parameters}
\label{tab:parameters}
\begin{tabularx}{\columnwidth}{@{}l l @{}}
\toprule
Parameter & Value \\
\midrule
Treadmill size l$\times$b$\times$h & 1744$\times$640$\times$520\,mm\\
Treadmill surface l$\times$ b & 1522$\times$510\,mm\\ 
Height from ground & 520\,mm\\ 
Total weight & 270\,kg\\ 
Floating weight & 45\, kg\\ 
Max speed & 4.7\,$\frac{m}{s} \approx$ 17\,$\frac{km}{h}$\\ 
Max subject weight & 500\,kg\\
Height handrail & 1100\,mm\\ 
Max. force & $F_x$, $F_y$ = 25\,kN\\ 
& $F_z$ = 60\,kN\\ 
Max. sampling frequency & 10\,000\,Hz \\ 
Safety features & Emergency stop button and safety key \\ 
& Handrail at 1100 \,mm height \\ 
& Adjustable lanyard with harness attached \\
& to the ceiling \\ 
\bottomrule
\end{tabularx}
\end{table}

\begin{figure}
	\centering
  \includegraphics[width=\columnwidth]{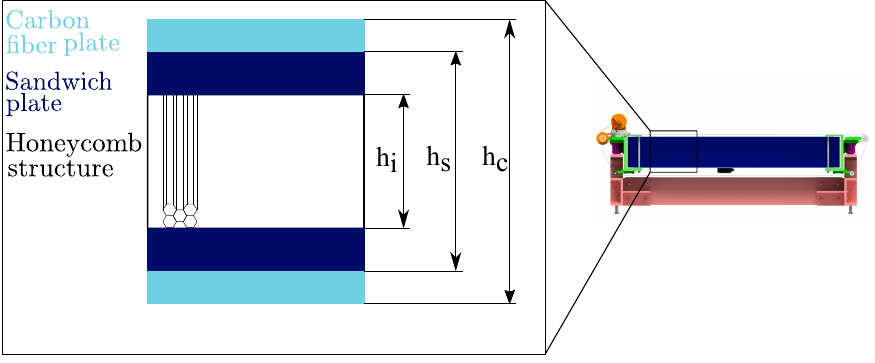}
	\caption{Sketch of a longitudinal cut through the sandwich (black) and carbon fiber (blue) compound treadmill plate.}
	\label{fig:thicknessSketch}
\end{figure}
\begin{figure}
	\centering
  \includegraphics[width=.75\columnwidth]{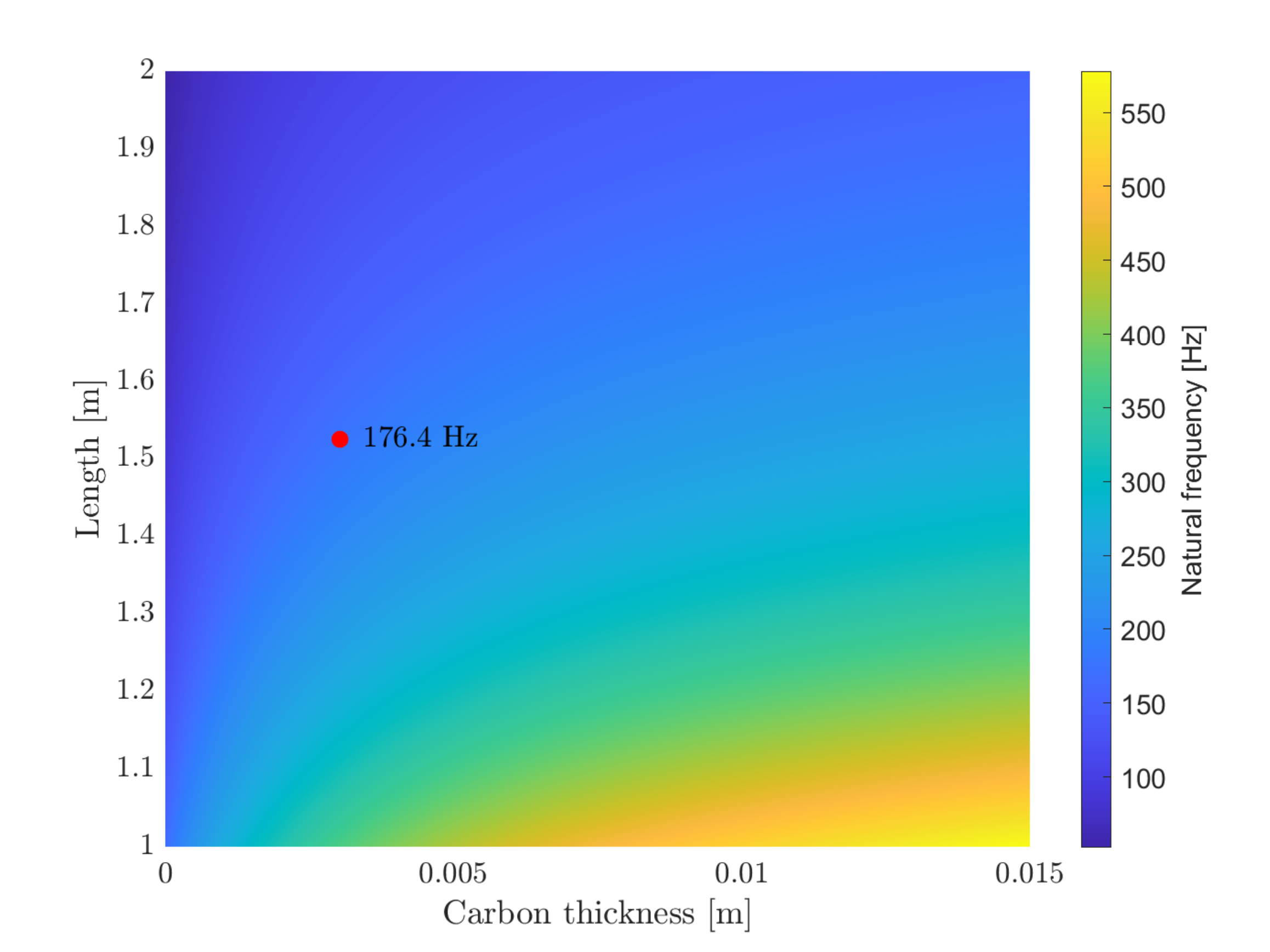}
	\caption{Heat map of the effects of treadmill plate length and carbon fiber plate thickness on the natural frequency of the instrumented treadmill. The red dot marks the selected value pair.}
	\label{fig:plateThicknessPlot}
\end{figure} 

To ensure safe experiments, a handrail is placed around the treadmill at 1100\,mm height according to DIN 18065. The handrail constructed from aluminium profiles (\textit{Bosch})  ensures safe experiments in the event of an accident. To not compromise the natural frequency of the treadmill, the handrail is not connected to the treadmill but rather placed on the ground. The handrail has cantilevers on the ground to prevent it from tipping over sideways.
The treadmill has a mobile control unit with a display for the current speed, a potentiometer to set the desired speed and an emergency stop button. On the handrail, there is also a safety key. If the subject moves too far from the front of the treadmill, the safety key attached to the subject is pulled from its base on the handrail and breaks the safety circuit. Both emergency stops are connected to the "Safe torque off" feature on the motor driver that enables safe error handling and immediately turns off the motor in case of an emergency.\\
Additionally, an adjustable lanyard (FAR1104-51 ,\textit{JSP}) is attached to the ceiling above the treadmill to prevent injuries in case an experiment participant stumbles on the treadmill. A wooden runway with a step is placed behind the treadmill for easier access as well as a safe area should a subject reach the end of the treadmill.\\
For synchronization, the charge amplifier, the high-speed camera and the treadmill motor controller are equipped with a custom-made trigger device.
\subsection{Theoretical Design Considerations}

The primary design criterion for instrumented treadmills is the natural frequency of the system. The higher the natural frequency, the higher the sensor accuracy (\autoref{tab:accuracy}). Under load, the treadmill will oscillate at its natural frequency. To not degrade the sensor signal, the natural frequency has to be higher than the frequency spectrum of the measured signal. The higher the natural frequency of the measuring systems is compared to the sensor signal spectrum, the cleaner the measured data will be. \\
In this section, we provide a simple model to estimate the natural frequency of our instrumented treadmill design for different design parameters. This way, the design is modular and can easily be modified for individual applications and desired dimensions. The maximum measurable frequency with a certain accuracy is defined by the force sensor. In this case, a piezoelectric force sensor (9366CCSP, \textit{Kistler},\cite{KistlerSensor}) defines the accuracy as a function of frequency based on the natural frequency $f_n$ of the measuring setup (\autoref{tab:accuracy}).
\begin{table}
\renewcommand{\arraystretch}{1.3}
\caption{Sensor accuracy as a function of frequency based on natural frequency $f_n$ \cite[p. 47]{KistlerSensor}}
\label{tab:accuracy}
\centering
\begin{tabularx}{0.8\columnwidth}{@{}X X X X @{}}
\toprule
Accuracy [\%]& 10 & 5 & 1\\
\midrule
$f_{max}$ & $0.3\cdot f_{n}$& $0.2\cdot f_{n}$& $0.1\cdot f_{n}$\\
\bottomrule
\end{tabularx}
\end{table}
The natural frequency is defined as:
\begin{equation}
f_n = \frac{1}{2\cdot\pi}\sqrt{\frac{c}{m}}
\end{equation}
where $m$ is the system mass and $c$ is the system stiffness. The part of the treadmill that has the biggest effect on the natural frequency is the plate the subject is running on (\autoref{fig:instrumentedTreadmillCut}). The design process is a trade-off between increasing stiffness, which is beneficial and increasing mass, which is disadvantageous. The goal is to find a material with high bending stiffness and low weight. To increase the stiffness of the plate \cite{Bundle2015} implemented honeycomb core sandwich panels (Compocel EL, \textit{CEL components}). Two aluminium plates provide a high area moment of inertia, while the hollow honeycomb structure provides rigidity with little increase to the system mass. We adopt the same principle but combine it with an epoxy-bonded compound carbon fiber plate on top of the aluminium plates. The carbon fiber plates increase the tensile strength on the surface of the treadmill plate to increase bending stiffness. At the same time, the lower density of carbon fiber adds less weight to the system than a thicker aluminium plate.\\

The driving equation for bending stiffness and natural frequency is 
\begin{equation}
\sigma = \frac{F\cdot l^3}{48\cdot E\cdot I}
\end{equation}
where $\sigma$ is the deformation, $F$ is the applied force, $E$ is the E-modulus and $I$ is the area moment of inertia. With Hooke's law 
\begin{equation}
c=\frac{F}{\sigma}
\end{equation}  this results in:
\begin{eqnarray}
f_n &=& \frac{1}{2\cdot \pi}\sqrt{\frac{48\cdot E\cdot I}{m\cdot l^3}}\\
I_{s} &=&\frac{b\cdot (h_{s}^3- h_{i}^3)}{ 12}\\
I_{c} &=&  \frac{b\cdot (h_{c}^3- h_{s}^3)}{ 12}\\
f_n &=& \sqrt{\frac{ b\cdot (E_s\cdot(h_{s}^3- h_{i}^3)+E_c\cdot(h_{c}^3- h_{s}^3))}{\pi^2 \cdot m\cdot l^3}}
\end{eqnarray}
where $I_s$ is the area moment of inertia of the sandwich plate, $I_c$ is the area moment of inertia of the carbon fiber plates, $E_s$ is the E-modulus of aluminium, $E_c$ is the E-modulus of a carbon fiber plate, b is the width of the treadmill surface, l is the length of the treadmill surface, $h_s$ is the thickness of the sandwich plate, $h_i$ is the thickness of the inside of the sandwich plate, and $h_c$ is the thickness of the sandwich carbon fiber compound plate as shown in \autoref{fig:thicknessSketch}. From the equation, we see that the stiffness of the treadmill is dominantly dependent on the length and height of the substrate.  
With a longer treadmill substrate and thinner carbon fiber plates, the natural frequency reduces as shown in \autoref{fig:plateThicknessPlot}. To find a good balance between low mass and high stiffness, we chose a treadmill area of $\approx 1400\times 500$\,mm. With these measures, we expect a natural frequency of $f_x=630$\,Hz, $f_y=250$\,Hz, and $f_z =176$\,Hz with minimal cost of material and weight. This natural frequency corresponds to the natural frequency of the gold standard force plate usually used for biomechanical experiments \cite{KistlerForceplate}.

\subsection{Calibration}
While the force sensors in the treadmill are factory calibrated, the treadmill has to be recalibrated after assembly to ensure the required natural frequency and accuracy are achieved. 
\subsubsection{Natural frequency}
To test the natural frequency, an impact test is used as described in the literature with either a wooden ball \cite{Paolini2007} or an instrumented hammer \cite{Sloot2015}. We tap the treadmill in all three Cartesian coordinate directions with a hammer while the data acquisition system is running. We measure the transient response to the impact, sampled at 10\,kHz, and use fast Fourier transform (FFT) to visualize the frequency response spectrum. The natural frequency $f_0$ for each direction shows as the prominent peak in the frequency response spectrum.\\
To determine the damping rate of the system, we calculate the envelope function of the transient response by fitting an exponential function to the peaks of each impact response. From the envelope function, we calculate the damping ratio $\xi$ and the Q-factor for each force direction.
\begin{align}
y(t) &= A\cdot e^{-\lambda\cdot t}\cdot \cos(\omega t)\\
\omega &= 2\pi\cdot f_0\\
\xi &= \frac{\lambda}{\sqrt{(\lambda^2+\omega^2)}}\\
Q &= \frac{1}{2\cdot \xi}
\end{align}
where $y(t)$ is the transient response, $A$ is the amplitude, $\lambda$ is the time constant, $t$ is time, $\omega$ is the oscillation angular velocity, $\xi$ is the damping ratio and $Q$ is the Q-factor.
\subsubsection{Signal noise}
To determine the motor-induced noise on the force sensors, we first collect force data when the motor is turned off and compare it to when the motor is running at speeds of 0-2.5\,$\frac{m}{s}$ in increments of 0.5\,$\frac{m}{s}$. We log the motor speed with the motor driver software at 1\,kHz and force data at 1\,kHz. We use two standard deviations (95\%) as a measure for the noise of the signal and display the changes in motor noise dependent on motor speed.\\
\subsubsection{Sensor calibration}
\begin{figure}
	\centering
  \includegraphics[width=.75\columnwidth]{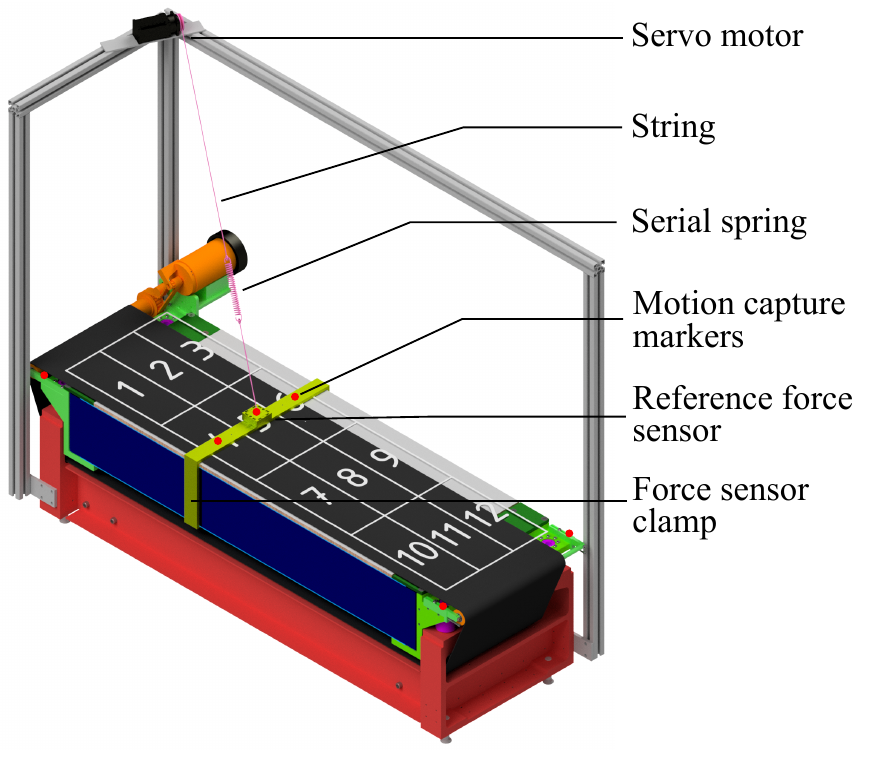}
  \caption{Calibration setup with treadmill, servo motor and force sensor for reproducible and accurate force insertion into the treadmill. The numbered grid represents the 12 calibration points. To measure orientation and position of the force sensor the treadmill, force sensor and force sensor clamp are equipped with motion capture markers.}
	\label{fig:calibrationSetup}
\end{figure}

\begin{figure}
	\centering
  \includegraphics[width=.75\columnwidth]{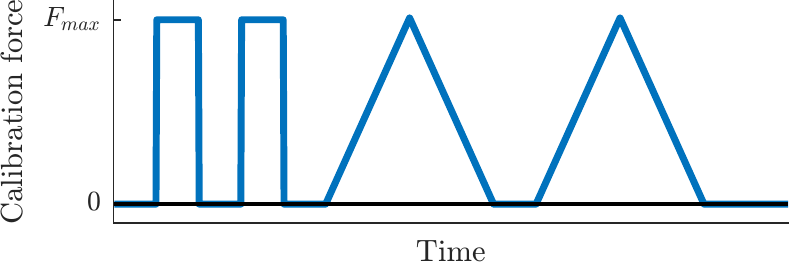}
	\caption{Calibration routine for the instrumented treadmill based on DKD 3-9 and ISO 376. The sensors are preloaded twice, and the force is then ramped up and down continuously twice. From this protocol, accuracy, linearity, repeatability, and hysteresis can be calculated.}
	\label{fig:calibrationRoutine}
\end{figure}

According to ISO 376 and DKD 3-9 (\autoref{fig:calibrationRoutine}) we calibrate the treadmill with a calibrated standard. 
To calibrate the treadmill, we use an external 3-axis force sensor (K3D60, \textit{ME-Systeme}) and a data acquisition system (cDAQ9189, \textit{National Instruments}). Using an external, calibrated force sensor as ground truth, we achieve a norm-conform calibration with more reliable data than a custom-built device.  To make the calibration reliable and reproducible, we use a servo motor (PM54-060-S250-R, \textit{Dynamixel}) to exert forces. The motor is fixed to the handrail in different positions. To connect the motor to the external force sensor, we use a 3\,mm Dyneema cable (Dyneema, \textit{Liros}) as well as a serial spring to provide series elastic behavior \cite{Pratt1995} to protect the external force sensor from shock. The external force sensor is clamped to the top of the treadmill in a grid of 4$\times$3 predefined positions (\autoref{fig:calibrationSetup}). To determine the force injection point, the force sensor position and orientation are recorded with a motion capture (MoCap) system (Prime$^{22}$, \textit{OptiTrack}) at 200\,Hz. Known fixpoints on the treadmill, the force sensor and the force sensor plate are equipped with MoCap markers (\autoref{fig:calibrationSetup}). This way, the exact orientation of the force sensor and the force injection points can be recalculated. The motor first exerts maximum force twice for 30\,s to preload the sensors. After the force is increased from 0\,N to 200\,N continuously by 6\,$\frac{N}{s}$. The force plate data is sampled at 1000\,Hz, the force sensor is sampled at 1600\,Hz and the motion capture data is recorded at 200\,Hz.\\
\begin{figure}
	\centering
  \includegraphics[width=.75\columnwidth]{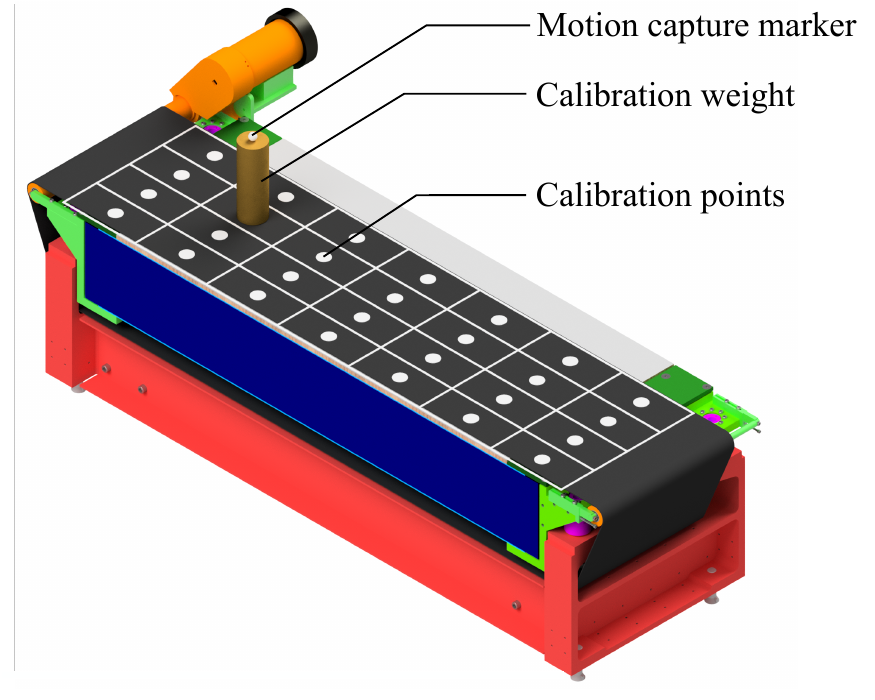}
  \caption{Setup for calibrating the COP accuracy. On the treadmill surface, a weight is placed on 28 points in an evenly spaced grid.  The weight is located with a MoCap marker, and the MoCap position is used as the ground truth to characterize the accuracy of the treadmill COP estimation.}
	\label{fig:copCalibrationSetup}
\end{figure}

\subsubsection{Shear compensation}
\begin{figure}
	\centering
  \includegraphics[width=.75\columnwidth]{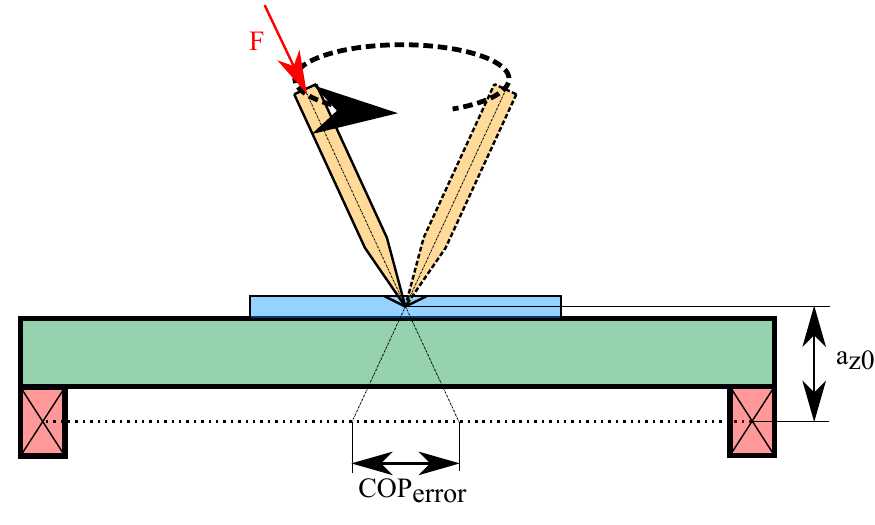}
	\caption{Sketch of the shear calibration procedure. A pointed rod (orange) induces a force F in different directions into the treadmill surface (green) through a conic hole in an aluminium plate (blue). Depending on the sensor offset $a_{z0}$, the influence of shear force components on the COP error in the sensor plane (red) can be minimized.}
	\label{fig:shearCalibrationSketch}
\end{figure}
Because the treadmill surface is not in one plane with the force sensor reference plane, shear effects on the center of pressure must be compensated (\autoref{fig:shearCalibrationSketch}). Therefore, we calculate an optimal sensor offset which is applied in the COP calculation to minimize the shear influence on the COP position. The equation results from the moment equilibrium around the treadmill centers:
\begin{eqnarray}
a_x = \frac{F_x*a_{z}-M_y}{F_z}\\
a_y = \frac{F_y*a_{z}+M_x}{F_z}
\end{eqnarray}
where $a_i$ is the COP position, $F_i$ are forces, $M_i$ are moments in Cartesian coordinates. $a_{z}$<0 is the sensor offset. We fix a thin aluminium plate with a conic hole to the treadmill surface and push into the whole with a pointy rod. This way, the force direction can change while ensuring that the force application position does not change. We perform 15 arbitrary motions with different force angles and amplitudes. Because the COP does not depend on the force amplitude, manual force application is acceptable. We then calculate the optimal $a_{zO}$ value that minimizes the change in $a_{x}$ and $a_y$ when the force direction changes. Finally, we find the optimal $a_{zO}$ value by subtracting the aluminium plate thickness.

\subsubsection{Center of pressure}

To estimate the COP error, we calibrate the COP position using a weight and a motion capture system (Mocap). We place a 15\,kg weight at 28 even-spaced points on the treadmill. The weight is marked at its center with a reflective Mocap marker. Using 3 Mocap cameras, we locate the weight on the treadmill. We use the Mocap data as ground truth to compare to the treadmill COP position estimate. The error of the treadmill estimate is calculated with:
\begin{equation}
COP_{error}= COP_{treadmill}-COP_{Mocap}
\end{equation}
We then fit a 4th order surface polynomial $S(x,y)$ to the $COP_{error}$ in x and y direction. With this polynomial, we correct future COP measurements to minimize the estimation error.
\begin{equation}
S(x,y) =  \sum_{n=0}^4 a_n \cdot x^n + b_n \cdot y^n
\end{equation}
where $S(x,y)$ is the surface polynomial, $a_n$ and $b_n$ are the polynomial coefficients and $x$ and $y$ are the Cartesian treadmill coordinates. The COP estimate for each treadmill point ($x_0$, $y_0$) can then be corrected with:
\begin{equation}
COP_{corrected_{(x_0,y_0)}} = COP_{Raw_{(x_0,y_0)}} - S(x_0,y_0)
\end{equation}
\subsection{Human and Robot Experimentation}
The switchable amplification of the force sensor signals (\autoref{tab:ranges}) in combination with the high natural frequency of the treadmill allows force measurements with constantly high accuracy independent of the force amplitude. To showcase the high accuracy of the treadmill for different loading scenarios, we showcase a human of 90\,kg and a small robot of 4\,kg, walking on the robot. Ground reaction force data is recorded for both subjects at 1\,kHz, and high-speed video of the legs is recorded at 400\,Hz. The treadmill speed is set to 0.63\,$\frac{m}{s}$.\\
The human subject walks on the treadmill for 1 minute, where 20\,s are recorded. The data is then normalized over body mass and is presented unfiltered and averaged for 30 strides. \\
We use our quadrupedal robot Morti \cite{Ruppert2021,ruppert_learning_2022} as comparison data for small subjects.  The robot walks on the treadmill, guided by a linear rail in the sagittal plane for 2 minutes, where 20\,s are recorded at a speed of 0.15\,$\frac{m}{s}$. We normalize the data by body weight and display it over 
the percent of step cycle.
\begin{figure*}
	\centering
  \includegraphics[width=.75\textwidth]{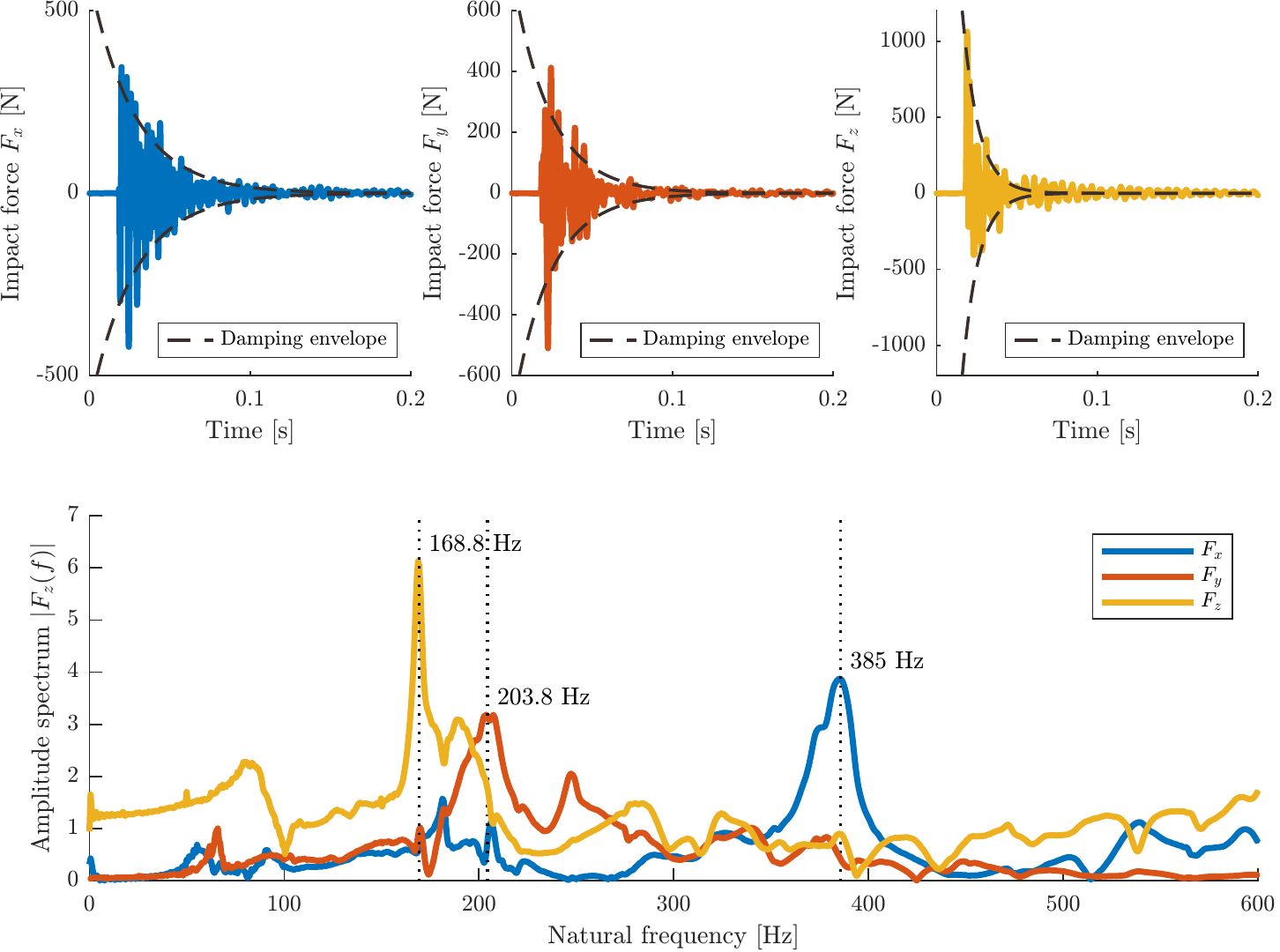}
	\caption{Unfiltered impact response of tapping the treadmill with a hammer in x-, y- and z-direction (top). Frequency response spectrum from Fast Fourier Transform (bottom). $F_x$ shows a prominent peek at 385\,Hz, $F_y$ at 203\,Hz and $F_z$ at 169\,Hz. These values correspond to the values calculated at the beginning of the paper.\\
The damping envelope (dashed line) is used to characterize the damping behavior. The damping ratios are $\xi_\mathrm{x} = 0.0148$, $\xi_\mathrm{y} = 0.033$ and $\xi_\mathrm{z}=0.0093$. The respective Q-factors are $Q_\mathrm{x}=33.8$, $Q_\mathrm{y}=15.02$ and $Q_\mathrm{z}=5.3$.}
	\label{fig:treadmillFrequency}
\end{figure*}
\section{Results}
 
\subsection{Calibration}
\subsubsection{Natural frequency}

The resulting natural frequencies (\autoref{fig:treadmillFrequency}) fit the theoretical values calculated in \autoref{fig:plateThicknessPlot}. The natural frequency in x-direction is the highest with $f_{0,x} = 385$\,Hz (calculated 635\,Hz). The deviation from the calculated frequency stems from the calculation being only a rough estimation since the standard equations for bending moment of inertia are mere approximations. The values for $f_{0,y} = 204$\,Hz (calculated 350\,Hz) and $f_{0,z} = 169$\,Hz (calculated 175\,Hz) fit well to the calculated values. As expected, the value for $f_{0,z}$ is the lowest since the stiffness is lowest in this direction, as discussed above. The deviations stem from uncertainty in material, manufacturing and assembly accuracy. All three spatial force directions show minor parasitic peaks in the frequency spectrum. Most noticeable is a parasitic peak in the z-direction at $\approx$ 80\,Hz. The parasitic peaks in the frequency spectrum stem from non-rigid connections on the drums and the actuator that oscillate at their natural frequencies. The prominent natural frequency peaks of the first harmonic of the treadmill structures are clearly visible at the frequencies mentioned above.\\
We also see the overall system damping in the time the oscillations fade out. In all three directions, the impact force fades after around 0.08\,s (\autoref{fig:treadmillFrequency} a,b,c).\\
To characterize the signal damping, we calculate the damping ratio $\xi$ and the Q-factor. 
The damping ratios are $\xi_\mathrm{x} = 0.0148$, $\xi_\mathrm{y} = 0.033$ and $\xi_\mathrm{z}=0.0093$. The respective Q-factors are $Q_\mathrm{x}=33.8$, $Q_\mathrm{y}=15.02$ and $Q_\mathrm{z}=5.3$. All damping rates are below 0.033, which shows that the construction is stiff and has few soft parts that interfere with the signal transfer through the compound plate.
\subsubsection{Signal noise}
\begin{figure}
	\centering
  \includegraphics[width=.75\columnwidth]{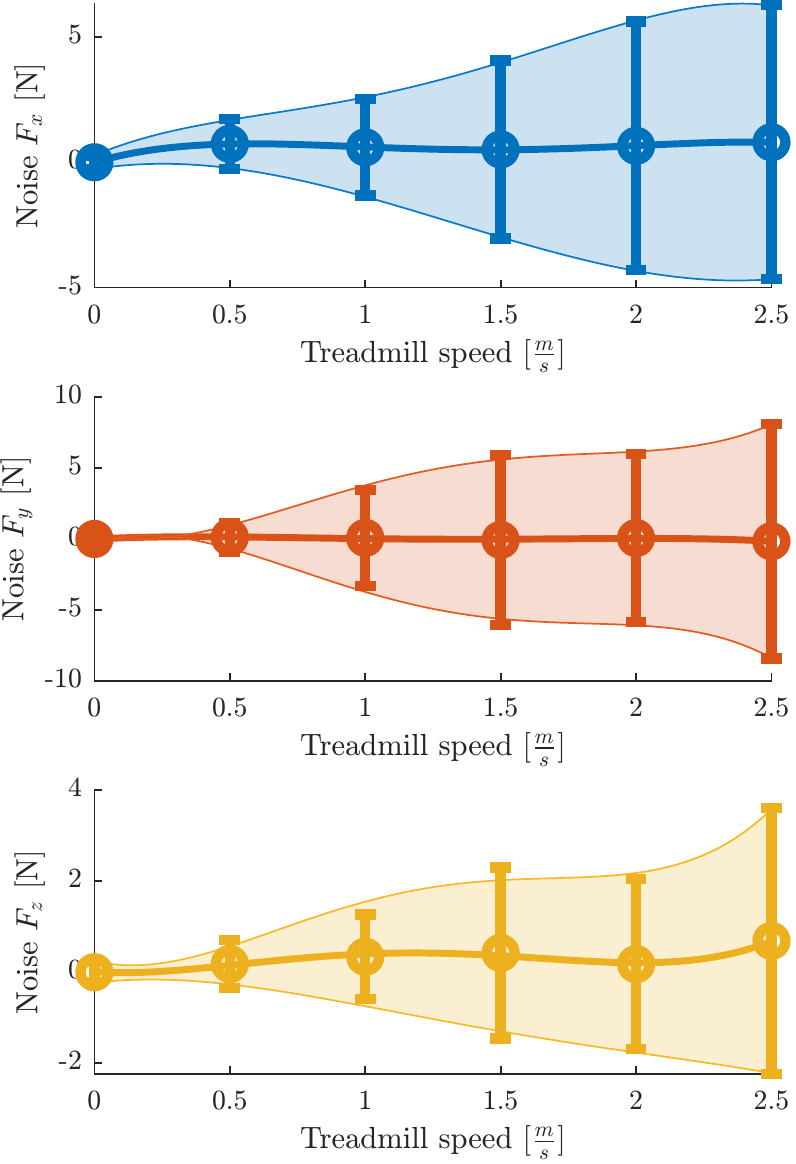}
	\caption{Noise level of the instrumented treadmill in relation to treadmill speed in x-, y- and z-direction with mean and 95\% confidence intervals. The frequency spectrum of the treadmill noise (bottom) shows prominent peaks at 50\,Hz in all three signals from line voltage.}
	\label{fig:zeroSpeed}
\end{figure}
The noise behavior of the treadmill at different treadmill speeds is shown in \autoref{fig:zeroSpeed}. The maximum noise of the force sensors at 0 $\frac{m}{s}$ is -0.76\,N, 0.77\,N, and -1.8\,N in x,y, and z-direction. The 95\% confidence interval in all three directions is $\leq$ 1\,N. At higher treadmill speeds, the noise level rises because of vibrations from the actuator. In x- and y-direction, the noise is higher with a maximum of 6 and 8\,N, respectively. Because the belt moves in the x-y plane, this is to be expected. In z-direction, the noise is $\leq$ 4\,N. Compared to the maximum range, as well as with respect to a human walking on the treadmill, the noise is negligible and would only come into play with very light subjects at high speeds.
\subsubsection{Sensor calibration}
\begin{figure}
	\centering
  \includegraphics[width=.75\columnwidth]{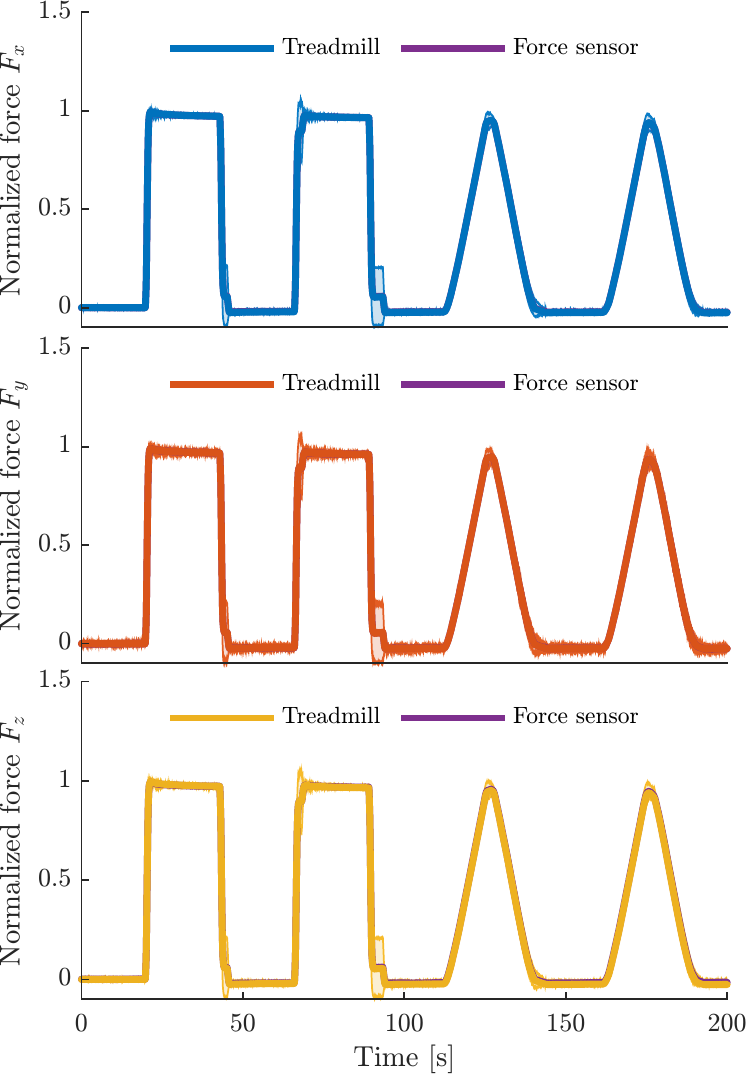}
	\caption{Averaged data over 12 calibration points from \autoref{fig:calibrationSetup} with treadmill data, 95\% confidence interval and ground truth force sensor data in purple. The calibration is done according to ISO 376 and DKD 3-9 as described in \autoref{fig:calibrationRoutine}.}
	\label{fig:calibrationExample}
\end{figure}
\begin{figure}
	\centering
  \includegraphics[width=\columnwidth]{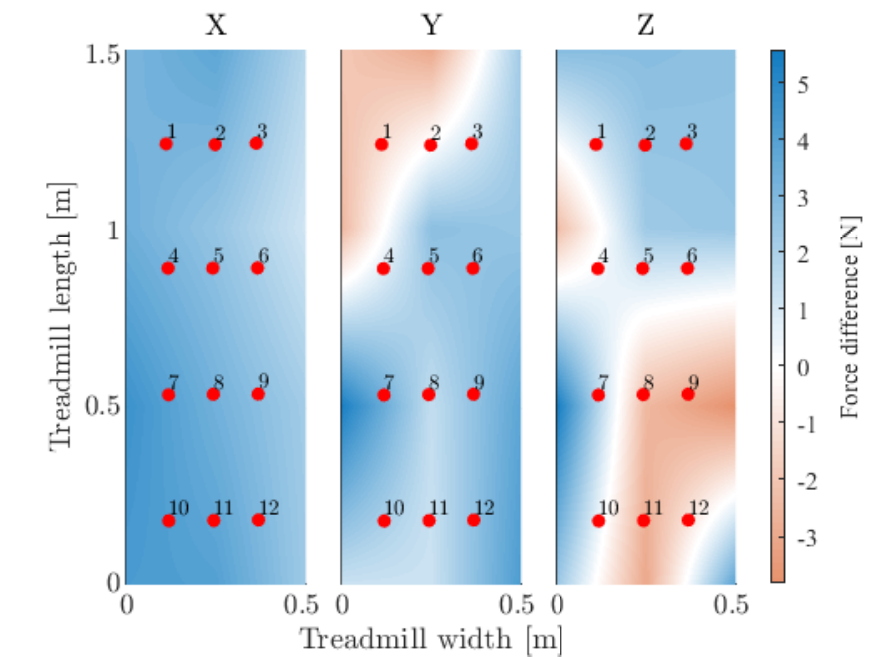}
	\caption{Calibration result for all calibration points shown in \autoref{fig:calibrationSetup}. Displayed is the maximum error between the force sensor and the treadmill data from \autoref{fig:calibrationExample}. The maximum error is 4.6\,N, 5.3\,N and 5.6\,N in x,y and z direction, respectively. The red dots show the 12 calibration points reconstructed from motion capture data.}
	\label{fig:staticTestError}
\end{figure}

The calibration data is compared to the force sensor data (\autoref{fig:calibrationExample}). For each calibration point, we calculate the maximum difference between the sensor as the reference and the treadmill. The sampled grid with the respective sensor accuracies in x, y and z direction is shown in \autoref{fig:staticTestError}. The maximum errors are $\Delta F_x = 4.6$\,N, $\Delta F_y = 5.3$\,N and $\Delta F_x = 5.6$\,N. Compared to their maximum sensor range, the error is around 0.02\%. The maximum error in our calibration occurs when the servo motor starts and stops moving rapidly at the discontinuities of the square functions. This induces oscillations in the serial spring, which lead to measuring errors. This can be omitted by using an elastic string to dampen the oscillation but is not necessary for our results since the calculations are only done at steady state behavior according to ISO 376. We, therefore, ignore the transient behaviors in between the evaluation windows.\\
The reproducibility was determined from the square function data. The difference between reoccurring sensor signals in \autoref{fig:calibrationExample} is 0.4\% of the normalized force. Linearity \cite{Czichosa} was determined from the difference of the data in the slope function to the linearized normal sensor data. It is $\leq$ 1\% for all three force directions. The hysteresis \cite{Czichosa} was calculated by fitting a 6th order polynomial to one of the slope functions, and the hysteresis effect was calculated in both input-output, as well as in output-input direction. The resulting hysteresis was found to be $\leq$ 0.25\% of the normalized force.
From all calibration points, we calculate a calibration map that shows the accuracy of the force-sensing capabilities for every point on the treadmill.
The calibration map is shown in \autoref{fig:staticTestError}. The error gets bigger towards the edges of the treadmill surface. This is to be expected because the bending of the sandwich plate is more asymmetric the further the force induction point deviates from the middle of the treadmill surface. Theoretically, force sensing is best in the center of the polygon spanned by the force sensors. This is especially visible in the x and y direction. A shift is also visible towards the left edge of the treadmill surface. We assume that the treadmill belt is not running at the exact center of the treadmill surface. Because the treadmill is designed as a split-belt treadmill, both belts are shifted toward one another to minimize the belt gap. The overall error in all three force directions lies well within our expected specifications for a treadmill of this size.

\subsubsection{Shear Compensation}
We calculate the minimal sensor offset from the 15 shear calibration experiments as described in \autoref{fig:shearCalibrationSketch}.
\begin{figure}
	\centering
  \includegraphics[width=.75\columnwidth]{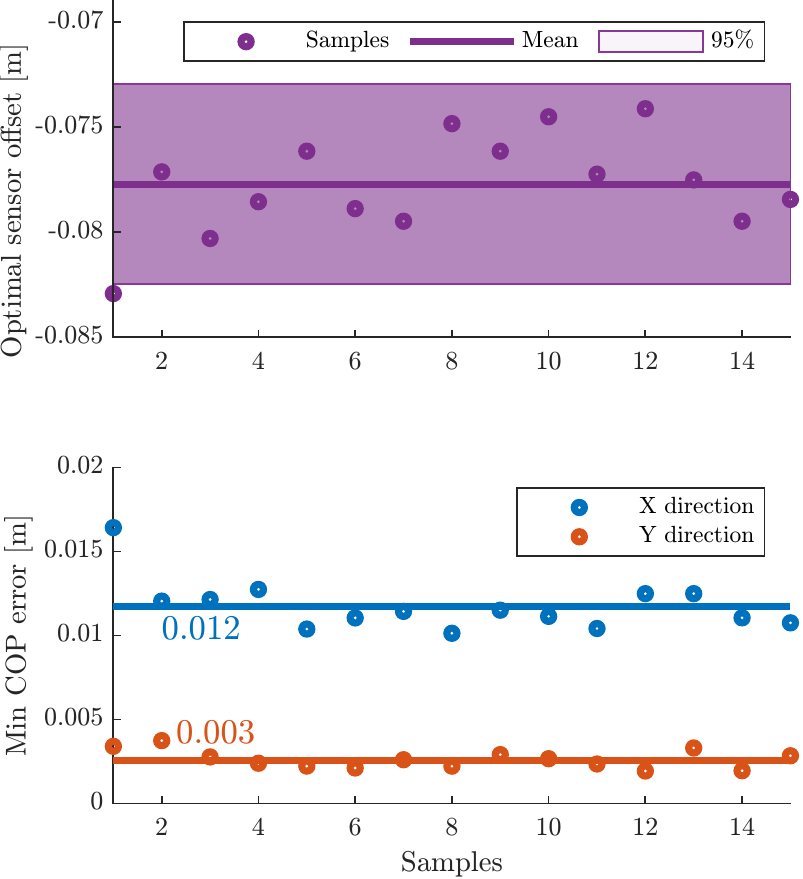}
	\caption{\textit{a)} Optimal sensor offset for 15 experiments with mean and 95\% confidence interval. The mean optimal sensor offset is -0.078\,m.\\
	\textit{b)} Resulting minimal COP estimate errors for the 15 experiments. Mean COP error from shear influence is 0.013\,m in x and 0.003\,m in y direction.}
	\label{fig:shearCalibration}
\end{figure}
The optimal sensor offset from 15 experiments is shown in \autoref{fig:shearCalibration} a). The mean sensor offset $a_{z,0}$ is 0.078\,m $\pm$0.005\,m. The deviation stems from the noise behavior characterized in \autoref{fig:zeroSpeed} as well as the measuring error in \autoref{fig:staticTestError}. With the optimized sensor distance, we recalculate the corrected COP from the sampled data. The resulting minimized expected COP error is 0.013 $\pm$0.0012\,m in x and 0.003$\pm$0.001\,m in y-direction as shown in \autoref{fig:shearCalibration} b). Therefore, the relative COP error is $\leq$ 0.8\% of the treadmill length in both directions. 

\subsubsection{Center of Pressure}
\begin{figure}
\centering
  \includegraphics[width=\columnwidth]{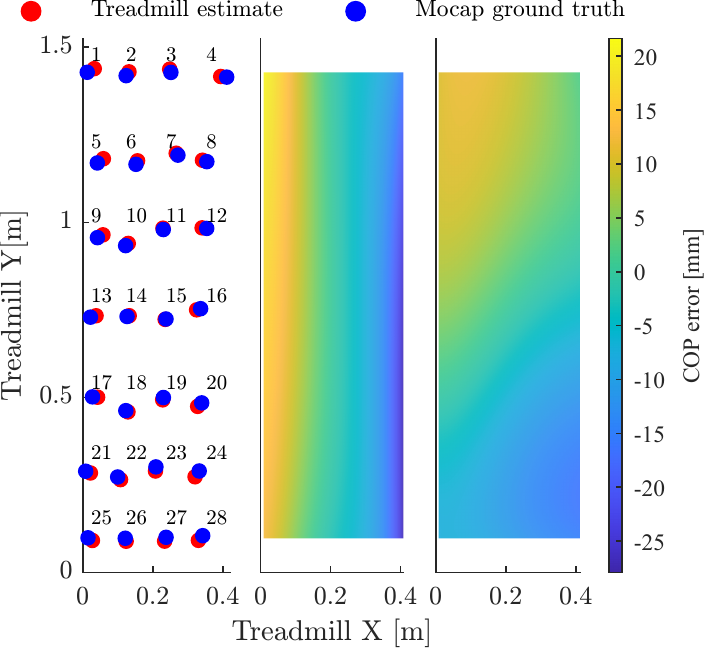}
\caption{\textit{a)} COP position from treadmill estimate (red) and ground truth (blue) from MoCap measurement. The raw COP error is $\leq 20$\,mm.
\textit{b)} COP error surface fit after COP calibration with a dead weight on 28 even-spaced calibration points across the treadmill surface, original data in orange.\\}
	\label{fig:COPgrid}
\end{figure}
The raw COP error in \autoref{fig:COPgrid} a) shows a COP estimation error of $\leq$ 20\,mm in x-direction and $\leq$ 12\,mm in y-direction. The fitted surface polynomial approximates the raw COP error data in \autoref{fig:COPgrid} b) well with $R^2$ = 99.86\%. With the COP error across the treadmill surface accounted for, the resulting error in the COP estimation is corrected with a remaining uncertainty of $\pm 1$\,mm in x direction and $\pm 0.5$\,mm in y direction. The error estimation through a surface polynomial shows to be an effective way of improving the COP estimate of the instrumented treadmill. 
\begin{figure}
	\centering
  \includegraphics[width=.75\columnwidth]{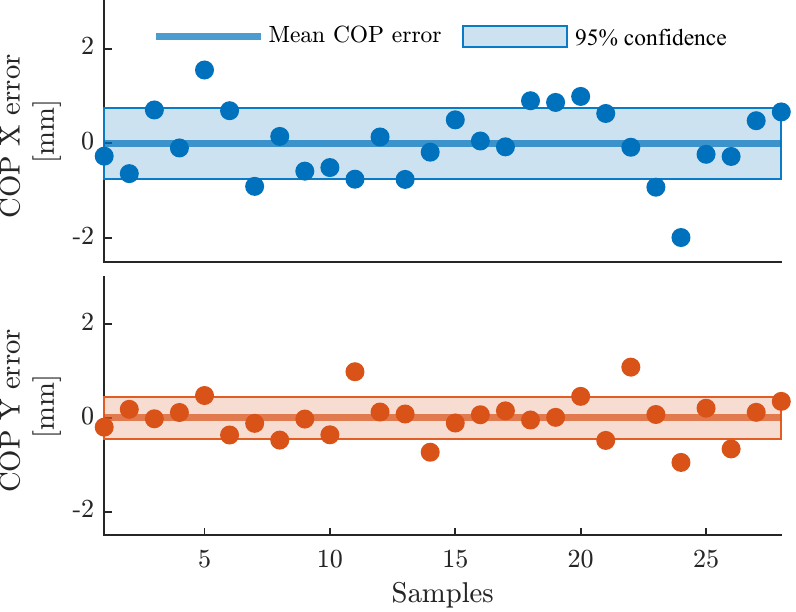}
	\caption{COP estimation after correction with the surface polynomial error approximation from \autoref{fig:COPgrid}. The COP error is corrected in both directions with a remaining uncertainty of $\pm 1$\,mm in x direction $\pm 0.5$\,mm in y direction.}
	\label{fig:copErrorCorrected}
\end{figure}
\subsection{Human and Robot Experimentation}
In the human walking experiment, we see an m-shaped ground reaction force profile expected for a walking gait. The z-direction shows one distinct maximum per step. We believe this to be due to the rigid structure of the treadmill as well as the subject wearing shoes \cite{Simon1981}. The overall shape is comparable to data presented in \cite{Kram1998} for a similar speed range.\\
To evaluate the performance of the instrumented treadmill, we calculate the confidence interval for the human and robot experiment over 30 steps. The confidence for the human treadmill are $S_{x,mean} = 8.8\pm 6.1$\,N, $S_{y,mean} = 8.4\pm 2.2$\,N and $S_{z,mean} = 16.8\pm 9.6$\,N. The confidence is highest during the maximum peak for each leg as well as during \\
The confidence for the robot experiment are $S_{x,mean} = 0.47\pm 0.19$\,N, $S_{y,mean} = 0.38\pm 0.16$\,N and $S_{z,mean} = 0.97\pm 0.44$\,N.\\
\begin{figure}
	\centering
  \includegraphics[width=.9\columnwidth]{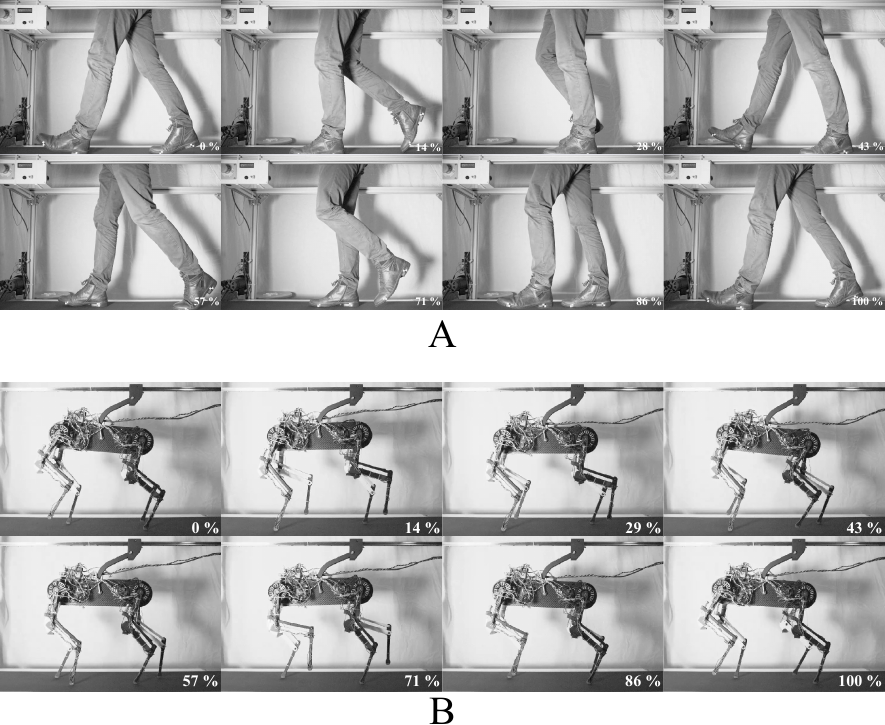}
	\caption{A. Photo of a human subject walking on the instrumented treadmill. B. Photo of a Robot walking on the instrumented treadmill.}
\label{fig:treadmillPhoto}
\end{figure}
\begin{figure}[b]
	\centering
	  \includegraphics[width=.75\columnwidth]{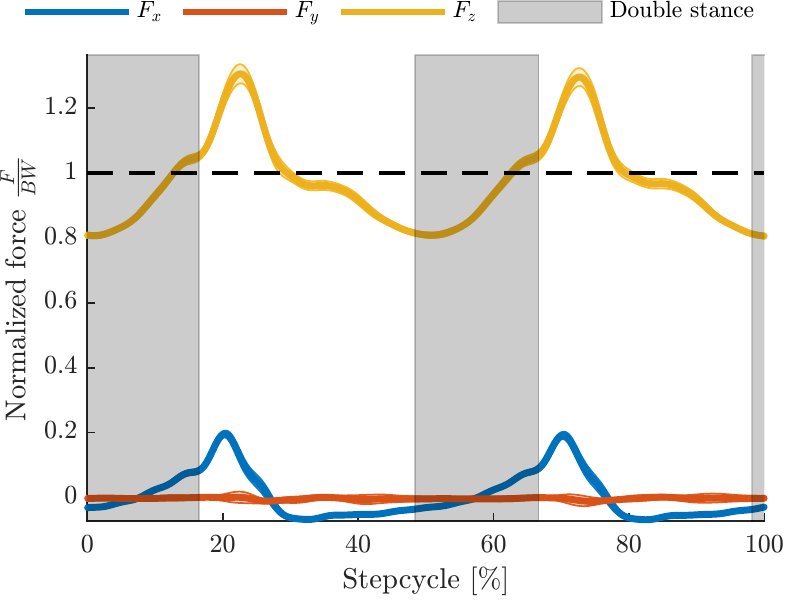}
	\caption{Normalized ground reaction force data for a human subject of 90\,kg walking on the treadmill at 1.3\,$\frac{m}{s}$. Data is shown for one stride consisting of one left and one right step. The double stance phase is shaded grey. All data is shown with 95\% confidence intervals, averaged over 30 consecutive steps in a 20\,s window. The body weight is shown as a dashed line.}
	\label{fig:humanData}
\end{figure}
\begin{figure}
	\centering
  \includegraphics[width=.75\columnwidth]{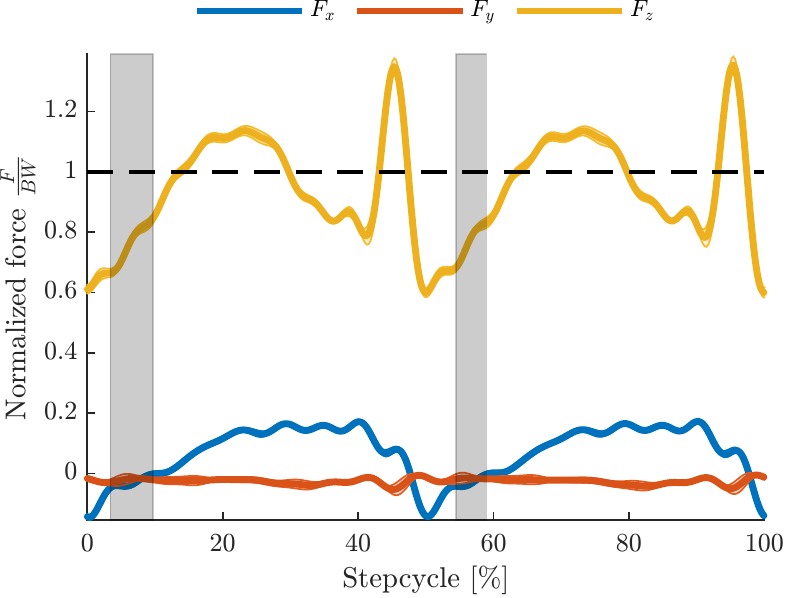}
	\caption{Normalized ground reaction force data for quadruped robot Morti of 4\,kg walking on the treadmill at 0.2\,$\frac{m}{s}$. Data is shown for one stride consisting of one front-left hind-right and one right-front hind-left step. The double stance phase is shaded grey. All data is shown with 95\% confidence intervals, averaged over 30 consecutive steps in a 20\,s window. The body weight is shown as a dashed line.}
	\label{fig:robotData}
\end{figure}
\section{Conclusion}
In this paper, we present a modular, instrumented treadmill. We provide a framework to design the natural frequency, the key feature, of such a treadmill. The framework can be used to design a treadmill of similar style with individual geometry and performance criteria. The sensor design in this paper is open-source and can be used as the basis for future development. We validate our design assumptions through experiments and show that, within manufacturing and material uncertainty, our design approach is valid. We reach a natural frequency of 169\,Hz which is less than 3\% off the calculated value. We then characterize the sensor performance of the treadmill with a norm conform calibration protocol according to ISO 376 and DKD 3-9. In the first step we calibrate the treadmill sensors against an external standard. This procedure provides a reproducible and accurate dataset that can then be used to calculate standard force sensor characteristics like accuracy, linearity, reproducibility and hysteresis. Since the treadmill is used as a sensor, we believe the proper calibration and report of characteristics are key features of an instrumented treadmill design. By providing a concise framework, we hope to improve the comparability and characterization procedure in the future. Overall the characterization of our treadmill shows values comparable to a standard force plate of that size. We then characterize the COP estimation by placing dead weight on the treadmill surface. With the surface fit correction provided by the force sensor manufacturer, we are able to achieve force plate levels of accuracy for the center of pressure estimation even though our treadmill surface is many folds bigger than a force plate. While the static COP error can be removed with minimal error, the influence of shear forces shows to be a much bigger influence. \\
We minimize the shear influence and the COP error can be decreased to 13$\times$3\,mm, which is sufficient for gait analyses. If required, the shear influence might be reduced by using appropriate filtering techniques and trying to estimate the force directions more accurately than in this approach.\\
Additionally, we characterize the noise behavior of the treadmill when the belt is actuated for different speeds. Overall the treadmill speed can be neglected and is only visible at high speeds.\\
Finally, we present a real-world experiment with a 90\,kg human as well as a 4\,kg robot to showcase the adjustable range and superior performance of our treadmill for different weight subjects.\\
The treadmill design presented here is comparable to the gold standard sensor modality for ground reaction forces: force plates. We achieve a similar natural frequency, good noise performance, and a center of pressure estimation well within the limits of what is typically required in biomechanical studies.\\
We achieve a lower floating mass through the compound plate design than other treadmill designs (\autoref{tab:treadmillComparison}). We achieve a much higher sampling frequency, better noise performance and accuracy. Through the norm conform calibration, we can calibrate with higher accuracy and reproduce our calibration results, which is not possible with other calibration concepts.
\section*{Contribution}
AAS and FR contributed to the concept, developed the mechanical design, the motor control and the experimental setup, conducted all experiments, analyzed the data and wrote the paper. ABS contributed to the concept, feedback, supervision and editing.
\section*{Availability}
The mechanical design and implementation details will be made publicly available upon this manuscript's acceptance. All data used in the paper is available upon request.
\section*{Acknowledgment}
We want to thank Bernard Javot for his feedback and help during the mechanical design, Andre Thill and Alessio Atzori for manufacturing and their help with assembly, and Maren Heuer for implementing the motor controller. We thank the International Max Planck Research School for Intelligent Systems (IMPRS-IS) for supporting the academic development of Felix Ruppert and Alborz Aghamaleki Sarvestani. This work was made possible thanks to a Max Planck Group Leader grant awarded to Alexander Badri-spr\"owitz by the Max Planck Society.
\printbibliography
\vspace{-3em}
\begin{IEEEbiography}[{\includegraphics[width=1in,keepaspectratio]{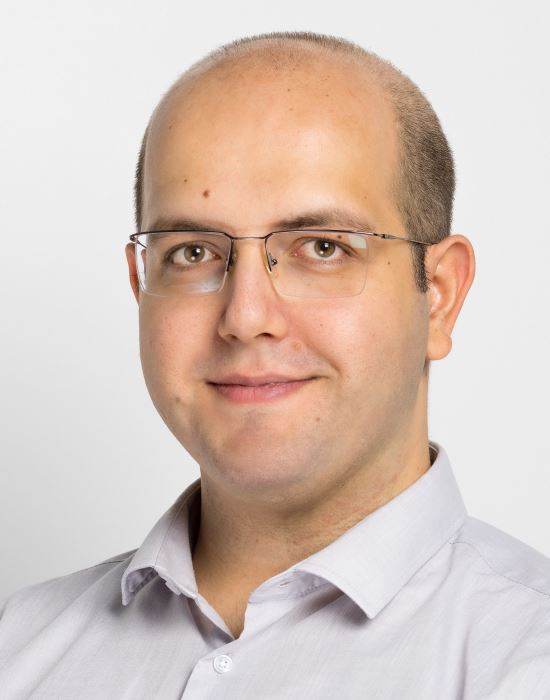}}]{Alborz Aghamaleki Sarvestani}
Received a B.Sc and M.Sc degree in mechanical engineering with a major in applied design, dynamics, vibrations, and control from Shiraz University in 2012 and 2015, respectively. He got his Ph.D. in robotics from the Max Planck Institute for Intelligent Systems. He is interested in legged locomotion, assistive robotics, bioinspired robotics, robotics, mechanism design, and sensor fusion.
\end{IEEEbiography}
\vspace{-3em}
\begin{IEEEbiography}[{\includegraphics[width=1in,keepaspectratio]{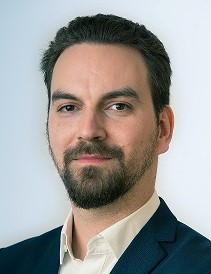}}]{Felix Ruppert}
received a B. Sc. and M. SC. in Mechatronics and Biomechatronics from Technische Universität Ilmenau and a Ph.D in robotics from the Max Planck Institute for Intelligent Systems. His research interests are the synergy of natural dynamics, neural control and learning in animal locomotion
\end{IEEEbiography}
\vspace{-3em}
\begin{IEEEbiography}[{\includegraphics[width=1in,keepaspectratio]{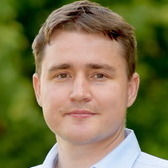}}]{Alexander Badri-Spr\"owitz}
received his Ph.D. in manufacturing systems and robotics from the Swiss Federal Institute of Technology, Lausanne, Switzerland in 2010. Since 2016, he leads the Max Planck Research Group Dynamic Locomotion at the MPI for Intelligent Systems in Stuttgart, Germany. In 2022 he joined the division Robotics, Automation and Mechatronics at KU Leuven in Belgium. His research focuses on mechanics, action, perception, and learning in animals and robots.
\end{IEEEbiography}
\end{document}